\documentclass[10pt,twocolumn,letterpaper]{article}

\usepackage{cvpr}              

\newcommand{\todo}[1]{{\color{red}#1}}

\definecolor{cvprblue}{rgb}{0.21,0.49,0.74}
\usepackage[pagebackref,breaklinks,colorlinks,citecolor=cvprblue]{hyperref}

\def\paperID{*****} 
\def\confName{CVPR}
\def\confYear{2026}

\title{The First Challenge on Mobile Real-World Image Super-Resolution at NTIRE 2026: Benchmark Results and Method Overview}

\author{
Jiatong Li
\thanks{\href{mailto:jiatong.li2024@gmail.com}{Jiatong Li},
\href{mailto:zhengchen.cse@gmail.com}{Zheng Chen},
\href{mailto:normal.kliu@gmail.com}{Kai Liu},
\href{mailto:jingkaiwang100@gmail.com}{Jingkai Wang}, 
\href{mailto:zzh_qwq@sjtu.edu.cn}{Zihan Zhou}, 
\href{mailto:liuxiaoyang0309@sjtu.edu.cn}{Xiaoyang Liu}, 
\href{mailto:labor030073@gmail.com}{Libo Zhu}, 
\href{mailto:g1017325431@gmail.com}{Jue Gong},  
\href{mailto:Radu.Timofte@uni-wuerzburg.de}{Radu Timofte}, 
and \href{mailto:yulun100@gmail.com}{Yulun Zhang} are the challenge organizers, while the other authors participated in the challenge. Section~B in the supplementary materials contains the authors' teams and affiliations. NTIRE 2026 webpage: \url{https://cvlai.net/ntire/2026}. Code: \url{https://github.com/jiatongli2024/NTIRE2026_Mobile_RealWorld_ImageSR}. }\and
Zheng Chen\footnotemark[1] \and
Kai Liu\footnotemark[1] \and
Jingkai Wang\footnotemark[1] \and
Zihan Zhou\footnotemark[1] \and
Xiaoyang Liu\footnotemark[1] \and
Libo Zhu\footnotemark[1] \and
Jue Gong\footnotemark[1] \and
Radu Timofte\footnotemark[1] \and
Yulun Zhang\footnotemark[1] \thanks{Corresponding author: Yulun Zhang. \href{mailto:yulun100@gmail.com}{yulun100@gmail.com}} \and
Congyu Wang \and
Zihao Wang \and
Ke Wu \and
Xinzhe Zhu \and
Fengkai Zhang \and
Zhongbao Yang \and
Long Sun \and
Jiangxin Dong \and
Jinshan Pan \and
Jiachen Tu \and
Yaokun Shi \and
Guoyi Xu \and
Yaoxin Jiang \and
Jiajia Liu \and
Renyuan Situ \and
Yixin Yang \and
Zhaorun Zhou \and
Junyang Chen \and
Yuqi Li \and
Chuanguang Yang \and
Weilun Feng \and
Chuanyue Yan \and
Yuedong Tan \and
Yingli Tian \and
Zhenzhong Chen \and
Tongqi Guo \and
Ruhan Liu \and
Sangzi Shi \and
Huazhang Deng \and
Jie Yang \and
Wenzhuo Ma \and
Yuantong Zhang \and
Daiqin Yang \and
Tianrun Chen \and
Deyi Ji \and
Yuxiao Jiang \and
Qi Zhu \and
Lanyun Zhu \and
Yuwen Pan \and
Runze Tian \and
Mingyu Shi \and
Zhanfeng Feng \and
Yuanfei Bao \and
Jiaming Guo \and
Renjing Pei  \and
Xin Di  \and
Long Peng \and
Linfeng Jiang \and
Xueyang Fu \and
Yang Cao \and
Zhengjun Zha \and
Choulhyouc Lee \and
Shyang-En Weng \and
Yi-Cheng Liao \and
Jorge Tyrakowski \and
Yu-Syuan Xu \and
Wei-Chen Chiu \and
Ching-Chun Huang \and
Yoonjin Im \and
Jihye Park \and
Hyungju Chun \and
Hyunhee Park \and
MinKyu Park \and
Xiaoxuan Yu \and
Jianxing Zhang \and
Yuxuan Jiang \and
Chengxi Zeng \and
Tianhao Peng \and
Fan Zhang \and
David Bull \and
Watchara Ruangsang \and
Supavadee Aramvith \and
JiaHao Deng \and
Wei Zhou \and
Hongyu Huang \and
Shaohui Lin \and
Zihan Wang \and
Yilin Chen \and
Yunchen Li \and
Junbo Qiao \and
Wei Li \and
Jiao Xie \and
Gaoqi He \and
Wenxi Li
}

\begin{document}

\maketitle
\begin{abstract}
This paper provides a review of the NTIRE 2026 challenge on mobile real-world image super-resolution, highlighting the proposed solutions and the resulting outcomes. The challenge aims to recover high-resolution (HR) images from low-resolution (LR) counterparts generated through unknown degradations with a $\times$4 scaling factor while ensuring the models remain executable on mobile devices. The objective is to develop effective and efficient network designs or solutions that achieve state-of-the-art real-world image super-resolution performance. The track of the challenge evaluates performance using a weighted combination of image quality assessment (IQA) score and speedup ratios. The competition attracted 108 registrants, with 16 teams achieving a valid score in the final ranking. This collaborative effort advances the performance of mobile real-world image super-resolution while offering an in-depth overview of the latest trends in the field.    
\end{abstract}
\section{Introduction}

\newcommand{\todobr}[1]{\todo{[TODO: #1]}}

Single image super-resolution (SR) aims to reconstruct a high-resolution (HR) image from its low-resolution (LR) counterpart, which is a fundamental and ill-posed inverse problem in computer vision. While SR has numerous applications in surveillance, medical imaging, and consumer photography, the practical deployment of SR models on mobile devices remains a significant challenge. Smartphones and edge platforms demand not only high perceptual quality but also low latency, small memory footprint, and energy efficiency—requirements that are often at odds with the ever-increasing size and complexity of modern SR networks.

Classical SR benchmarks typically assume a simple bicubic downsampling degradation, while real-world degradations (e.g., sensor noise, compression artifacts, unknown blur) are far more complex and rarely follow a simple bicubic model. Over the years, the field has evolved from interpolation-based methods~\cite{zibetti2007robust,sun2010context,zhang2012single} and shallow CNNs~\cite{dong2014learning,kim2016accurate,zhang2018image,wang2015deep,zhang2018residual} to deep Transformers~\cite{vaswani2017attention,liu2021swin,chen2023dual,dai2019second} and state-space models (Mamba)~\cite{gu2023mamba,liu2024vmamba,guo2024mambair}, achieving remarkable reconstruction accuracy. More recently, generative models such as GANs~\cite{goodfellow2014generative,zhang2017image,ledig2017photo} and diffusion models~\cite{ho2020denoising,saharia2022image,xia2023diffir,wu2024seesr} have pushed the boundaries of perceptual realism. 

However, these advanced models typically require billions of parameters and massive computational budgets, making them impractical for real-time on-device inference. For example, a diffusion-based SR model may take several seconds to process a single image on a mobile accelerator, while consuming hundreds of megabytes of memory. With the increasing demand for real-time image enhancement on smartphones, wearable devices, and edge platforms, it is essential to design SR solutions that are both perceptually realistic and sufficiently lightweight to run on resource-constrained hardware.

To bridge this gap between SR performance and mobile deployability, we organize the NTIRE 2026 Challenge on Mobile Real-World Image Super-Resolution, held in conjunction with the 11th NTIRE workshop at CVPR 2026. This challenge focuses on the real-world degradation setting (scaling factor \(\times4\)), with a strong emphasis on efficiency. The key distinguishing features are:

\begin{itemize}
    \item \textbf{Rigorous mobile constraints}: To ensure practical deployability, we impose strict limits on model parameters and inference FLOPs, and restrict supported operators to those commonly accelerated on mobile AI chips.
    \item \textbf{Joint optimization of perceptual quality and speed}: The evaluation combines perceptual metrics and relative inference speedup on the target hardware. A higher final score indicates a better trade-off, rewarding solutions that achieve both high visual quality and low latency.
\end{itemize}

This challenge is one of the challenges associated with the NTIRE 2026 Workshop~\footnote{\url{https://www.cvlai.net/ntire/2026/}} on:
deepfake detection~\cite{ntire26deepfake}, 
high-resolution depth~\cite{ntire26hrdepth},
multi-exposure image fusion~\cite{ntire26raim_fusion}, 
AI flash portrait~\cite{ntire26raim_portrait}, 
professional image quality assessment~\cite{ntire26raim_piqa},
light field super-resolution~\cite{ntire26lightsr},
3D content super-resolution~\cite{ntire263dsr},
bitstream-corrupted video restoration~\cite{ntire26videores},
X-AIGC quality assessment~\cite{ntire26XAIGCqa},
shadow removal~\cite{ntire26shadow},
ambient lighting normalization~\cite{ntire26lightnorm},
controllable Bokeh rendering~\cite{ntire26bokeh},
rip current detection and segmentation~\cite{ntire26ripdetseg},
low light image enhancement~\cite{ntire26llie},
high FPS video frame interpolation~\cite{ntire26highfps},
Night-time dehazing~\cite{ntire26nthaze,ntire26nthaze_rep},
learned ISP with unpaired data~\cite{ntire26isp},
short-form UGC video restoration~\cite{ntire26ugcvideo},
raindrop removal for dual-focused images~\cite{ntire26dual_focus},
image super-resolution (x4)~\cite{ntire26srx4},
photography retouching transfer~\cite{ntire26retouching},
mobile real-word super-resolution~\cite{ntire26rwsr},
remote sensing infrared super-resolution~\cite{ntire26rsirsr},
AI-Generated image detection~\cite{ntire26aigendet},
cross-domain few-shot object detection~\cite{ntire26cdfsod},
financial receipt restoration and reasoning~\cite{ntire26finrec},
real-world face restoration~\cite{ntire26faceres},
reflection removal~\cite{ntire26reflection},
anomaly detection of face enhancement~\cite{ntire26anomalydet},
video saliency prediction~\cite{ntire26videosal},
efficient super-resolution~\cite{ntire26effsr},
3d restoration and reconstruction in adverse conditions~\cite{ntire26realx3d},
image denoising~\cite{ntire26denoising},
blind computational aberration correction~\cite{ntire26aberration},
event-based image deblurring~\cite{ntire26eventblurr},
efficient burst HDR and restoration~\cite{ntire26bursthdr},
low-light enhancement: `twilight cowboy'~\cite{ntire26twilight},
and efficient low light image enhancement~\cite{ntire26effllie}.

\begin{table*}[t]
\setlength{\tabcolsep}{1.3mm}
\centering
\resizebox{\textwidth}{!}{
\begin{tabular}{c l | c | c c c c c c c | c c}
\toprule
\textbf{Team Number} & \textbf{Team Name} & \textbf{Rank} & \textbf{LPIPS} & \textbf{DISTS} & \textbf{NIQE} & \textbf{ManIQA} & \textbf{MUSIQ} & \textbf{CLIP-IQA} & \textbf{Speedup} & \textbf{Perceptual Score} & \textbf{Total Score} \\
\midrule
14 & VIPSL           & 1  & 0.3619 & 0.1931 & 3.474 & 0.3798 & 68.91 & 0.8669 & 113.85 & 4.0334 & 42.2125 \\
6  & Antman          & 2  & 0.3494 & 0.2048 & 5.419 & 0.6723 & 71.97 & 0.8700 & 21.79  & 4.1659 & 33.2446 \\
11 & SamsungAICamera & 3  & 0.3524 & 0.2232 & 2.810 & 0.4684 & 74.68 & 0.8805 & 11.67  & 4.2392 & 30.8701 \\
3  & TODSR           & 4  & 0.3256 & 0.1756 & 2.954 & 0.5450 & 73.58 & 0.8680 & 1.00   & 4.3522 & 20.4241 \\
15 & YuFans          & 5  & 0.4676 & 0.2172 & 3.525 & 0.3394 & 56.18 & 0.5964 & 21.79  & 3.4603 & 20.3844 \\
4  & IMAG2006        & 6  & 0.3649 & 0.2009 & 3.722 & 0.6523 & 73.47 & 0.8769 & 1.00   & 4.3259 & 20.0547 \\
8  & Super03         & 7  & 0.3288 & 0.1712 & 3.954 & 0.4172 & 67.78 & 0.6884 & 4.48   & 3.8881 & 19.9840 \\
1  & VEPG            & 8  & 0.3562 & 0.1800 & 3.710 & 0.5352 & 73.70 & 0.8788 & 1.00   & 4.2437 & 18.9444 \\
9  & SnowVision      & 9  & 0.4333 & 0.2424 & 6.396 & 0.2288 & 38.02 & 0.3760 & 143.35 & 2.6698 & 17.1779 \\
12 & BVISR           & 10 & 0.3358 & 0.1657 & 3.550 & 0.5079 & 70.30 & 0.7477 & 1.00   & 4.1021 & 17.1730 \\
16 & EIC-ECNU        & 11 & 0.2134 & 0.1083 & 3.263 & 0.4101 & 66.34 & 0.6452 & 1.00   & 4.0707 & 16.8032 \\
2  & NTR             & 12 & 0.3725 & 0.1758 & 3.803 & 0.4776 & 69.59 & 0.7686 & 1.00   & 4.0126 & 16.1404 \\
5  & NoReject        & 13 & 0.3214 & 0.1660 & 3.548 & 0.4178 & 69.30 & 0.7092 & 1.00   & 3.9777 & 15.7547 \\
10 & ACM\_HCC        & 14 & 0.3617 & 0.1780 & 3.612 & 0.4439 & 69.48 & 0.7063 & 0.99   & 3.9440 & 15.3601 \\
13 & MDAP            & 15 & 0.4543 & 0.2566 & 7.370 & 0.2270 & 29.51 & 0.3920 & 94.95  & 2.4662 & 13.7370 \\
7  & SFVision        & 16 & 0.5241 & 0.2889 & 8.490 & 0.2339 & 23.91 & 0.4197 & 10.21  & 2.2307 & 7.4696  \\
\bottomrule
\end{tabular}
}
\caption{\textbf{Results of NTIRE 2026 Mobile Real-World Image Super-Resolution Challenge.} Perceptual score is a weighted combination of six perceptual metric scores, and is evaluated on the DIV2K validation set (100 images). Rankings are determined by weighting the perceptual score and the speedup ratio relative to the OSEDiff~\cite{wu2024osediff} on the MediaTek Dimensity 8400 platform.
The Team descriptions in the main paper Sec.~\ref{sec:methods} and supplementary material Sec.~\textcolor{red}{A} are ordered according to this table.}
\label{tab:team_rankings}
\end{table*}

\section{NTIRE 2026 Challenge on Mobile Real-World Image Super-Resolution}

The NTIRE 2026 Challenge on Mobile Real-World Image Super-Resolution is one of the associated challenges of NTIRE 2026, pursuing two main goals. First, it seeks to advance the state of the art in real-world SR under unknown degradations while maintaining high efficiency for mobile deployment. Second, it provides a venue where researchers and industry practitioners can collaborate to develop practical solutions for on-device image enhancement. The sections below describe the specific details of the challenge.

\subsection{Dataset}

The challenge officially provides two datasets: DIV2K~\cite{agustsson2017div2k} and LSDIR~\cite{li2023lsdir}. Participants are also allowed to use extra external data for training. For both datasets, LR-HR pairs are constructed from high-quality images using a downsampling factor of $\times 4$.

\noindent\textbf{DIV2K}
DIV2K includes 1,000 images at 2K resolution, divided into 800 training images, 100 validation images, and 100 testing images. To ensure a fair comparison, the HR images of the DIV2K validation subset are hidden from participants until the testing stage. The HR images of the test subset remain undisclosed throughout the whole challenge.

\noindent\textbf{LSDIR}
The LSDIR dataset contains 86,991 images sourced from the Flickr platform, partitioned into three subsets of 84,991/1000/1000 for training, validation, and testing.

\subsection{Track and Competition}
\label{sec:evaluation}
The competition features a single track that balances perceptual quality and inference speed.

\noindent\textbf{Perceptual Quality Metrics.}
Six widely used IQA metrics are employed to comprehensively assess the restored results: LPIPS~\cite{zhang2018lpips}, DISTS~\cite{ding2020dists}, CLIP-IQA~\cite{wang2023clipiqa}, MANIQA~\cite{yang2022maniqa}, MUSIQ~\cite{ke2021musiq}, and NIQE~\cite{zhang2015niqe}. The overall perceptual score is defined as:

\begin{equation}
\begin{aligned}
    \text{Score} = \left(1 - \text{LPIPS}\right) + \left(1 - \text{DISTS}\right) + \text{CLIP-IQA} \\
    + \text{MANIQA} + \frac{\text{MUSIQ}}{100} + \max\left(0, \frac{10 - \text{NIQE}}{10}\right).
\end{aligned}
\end{equation}

\noindent\textbf{Inference Speed.}
The inference speed is measured on the MediaTek Dimensity 8400 platform with FP16 precision. The input image size is \(128\times128\) and the output size is \(512\times512\). Let \(t_{\text{osediff}}\) denote the average inference time of the baseline model OSEDiff~\cite{wu2024osediff} on the same platform, and \(t_{\text{curmodel}}\) denote the time of the submitted model. 

The speedup ratio is:

\begin{equation}
\begin{aligned}
    \text{Speedup} = \frac{t_{\text{osediff}}}{t_{\text{curmodel}}}.
\end{aligned}
\end{equation}

\paragraph{Final Score.}
The final score balances perceptual quality and speed:

\begin{equation}
\begin{aligned}
    \text{FinalScore} = 2^{\text{Score}} \cdot \text{Speedup}^{0.2}.
\end{aligned}
\end{equation}

\noindent\textbf{Challenge Phases.}
\textit{(1) Development and Validation Phase:} Participants are given access to two datasets: (a) 800 LR/HR training image pairs and 100 LR validation images from the DIV2K dataset, and (b) 84,991 LR/HR training image pairs and 1,000 LR validation images from the LSDIR dataset, and may use additional data. They can upload their restored HR images for the test LR images to the Codabench server, which returns the perceptual metrics computed against the hidden HR images, allowing refinement.

\textit{(2) Testing Phase:} 100 test LR images from DIV2K validation set are used for final evaluation. Participants are required to submit their SR outputs to the Codabench server, along with their code, model checkpoints, and a detailed report sent to the organizers via email. The final ranking is determined by the organizers after verifying the submitted code and measuring the inference speed on the MediaTek Dimensity 8400 platform.

\noindent\textbf{Evaluation Protocol.} 
All calculations are carried out on the Y channel of the YCbCr color space. The results shown on the Codabench leaderboard are for reference only; the final rankings will be based on the official evaluation reproducted with the submitted code on the MediaTek Dimensity 8400 platform and the provided test set. An evaluation script for these metrics is available at \url{https://github.com/jiatongli2024/NTIRE2026_Mobile_RealWorld_ImageSR}, which also contains the source code and pre-trained models.

\begin{figure*}[t]
\centering
\includegraphics[width=\linewidth]{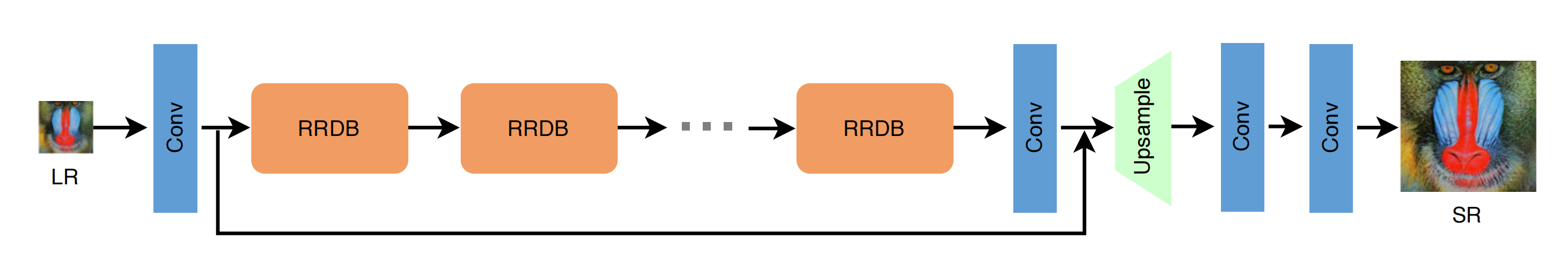}
\caption{\textbf{Team Antman}}
\label{fig:team06_structure}
\end{figure*}

\section{Challenge Results}
Table~\ref{tab:team_rankings} presents the final rankings and results of the teams.
A comprehensive description of the evaluation process is
outlined in Sec.~\ref{sec:evaluation}. The top 5 teams, along with their
method details, are provided in Sec.~\ref{sec:methods}, while the remaining teams are listed in Sec. A of the supplementary materials. In addition, team member information can be found
in Sec. B of the supplementary materials.

\subsection{Architectures and main ideas}
Participants explored diverse technical approaches to strike an optimal balance between super-resolution performance and computational efficiency. Below, we summarize the key innovations and contributions from the teams.

\begin{enumerate}
    \item \textbf{One-step diffusion models and LoRA adaptation dominate the foundation.}
    Rather than relying on traditional multi-step diffusion, one-step diffusion frameworks built upon Stable Diffusion ~\cite{rombach2022ldm} were heavily favored. Methods such as OSEDiff~\cite{wu2024osediff}, TADSR~\cite{zhang2025time}, PiSA-SR~\cite{sun2025pisasr} and OMGSR~\cite{wu2025omgsr} served as the backbone for many competitive solutions. The TODSR team, for example, adapted PiSA-SR~\cite{sun2025pisasr} to emphasize that latent alignment and semantic guidance should be jointly coordinated to avoid unstable over-enhancement.

    \vspace{2.5mm}
    \item \textbf{Knowledge distillation bridges the gap between massive generative priors and mobile efficiency.}
    To meet the strict parameter and FLOP constraints of mobile deployment, several teams employed distillation strategies to transfer knowledge from heavy teacher models to lightweight students. For example, the  SamsungAICamera team utilized adversarial diffusion compression with aggressively pruned UNets and lightweight decoders like TAESD~\cite{bohan2023taesd}, achieving competitive perceptual performance while significantly reducing the inference overhead of the diffusion model. 

    \vspace{2.5mm}
    \item \textbf{Advanced latent inversion and timestep alignment improve diffusion stability.}
    To better adapt diffusion processes for super-resolution, participants proposed sophisticated conditioning mechanisms. 
    These methods aim to anchor the low-quality input more accurately onto the diffusion trajectory, reducing hallucinated artifacts.
    The team TODSR, for example, introduced Latent-Timestep Alignment (LSA) to align low-quality latent features with diffusion latent states.

    \vspace{2.5mm}
    \item \textbf{Hybrid architectures and explicit local refinement modules preserve high-frequency details.}
    To resolve the trade-off between the structural consistency of GANs and the perceptual richness of latent diffusion models, teams fused different architectural paradigms. For example, team YuFans used a two-model ensemble approach that combines the complementary strengths of GAN-based and diffusion-based super-resolution to maximize perceptual quality metrics.

    \vspace{2.5mm}
    \item \textbf{Multi-stage and progressive fine-tuning curricula are standard.}
    Training strategies shifted from single-phase optimization toward carefully staged adaptation. Typical pipelines involved a base-adaptation phase focused on structural reconstruction or mild synthetic degradations, followed by secondary phases targeting complex real-world domain adaptation, semantic LoRA fine-tuning, or score-oriented perceptual refinement. The EIC-ECNU team, for example, proposed a hierarchical dual-LoRA structure to separate base restoration from complex artifact refinement to get better performance.

\end{enumerate}
\subsection{Participants}
This year, the mobile real-world image super-resolution challenge saw 108
registered participants, with 16 teams submitting their valid
models, and had a valid score in the final
ranking. These entries set a new standard for the state-of-the-art in mobile image super-resolution.

\subsection{Fairness}
A set of rules has been established to ensure the fairness of
the competition. (1) The use of DIV2K validation HR images for
training is strictly prohibited. (2) Participants are allowed to train
using additional datasets, such as the Flickr2K~\cite{lim2017enhanced} and LSDIR
datasets. (3) The application of
no-reference IQA and simulated degradation pipelines during training and testing is considered a fair practice.

\begin{figure*}[tbph]
\centering
\includegraphics[width=\linewidth]{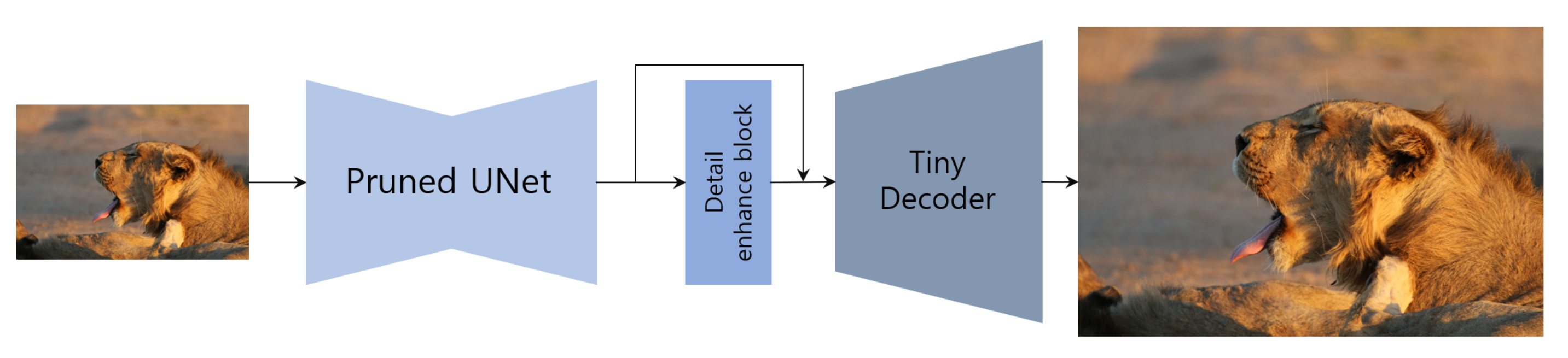}
\caption{\textbf{Team  SamsungAICamera}}
\label{fig:team11_structure}
\end{figure*}

\subsection{Conclusions}
The insights gained from analyzing the results of the NTIRE 2026 mobile real-world image super-resolution challenge are summarized as follows:
\begin{enumerate}
\item Methods leveraging priors from models like Stable Diffusion~\cite{rombach2022ldm}, particularly through one-step architectures like OSEDiff~\cite{wu2024osediff}, provide highly competitive perceptual realism and vivid texture.

\item Knowledge distillation and architectural compression are mandatory for mobile deployment. To bridge the gap between heavy generative priors and mobile hardware constraints, participants successfully utilized adversarial distillation, pseudo-ground-truth generation, and network pruning to transfer capabilities from heavy teacher models to lightweight student networks. 

\item Latent space optimization and timestep adaptation improve diffusion stability. Rather than relying on fixed timesteps or stochastic noise injection, advanced techniques such as deterministic inversion and latent-timestep alignment were introduced. These strategies help anchor the low-quality input properly onto the diffusion manifold, reducing structural distortions.

\item Hybrid paradigms and residual refinement resolve the perception-distortion trade-off. Combining the structural reliability of GANs with the texture generation of diffusion models emerged as a highly effective strategy. 

\end{enumerate}

\section{Challenge Methods and Teams}
\label{sec:methods}
\subsection{VIPSL}
\textbf{Description.}
VIPSL builds an efficient perceptual SR pipeline around PLKSR-Rep~\cite{lee2024plksr}, targeting the mobile real-world setting where both quality and inference efficiency matter. Instead of introducing a heavy architecture, the team keeps the backbone compact (dim $=64$, $12$ blocks) and focuses on a robust fine-tuning strategy. The training pipeline follows Real-ESRGAN-style degradation simulation~\cite{wang2021realesrgan} and BasicSR engineering practice~\cite{wang2022basicsr}, which provides stable behavior across diverse degradations.

The key idea is to use IQA-aware optimization to better align with the challenge scoring criteria. In addition to pixel-level fidelity, the method explicitly accounts for perceptual criteria through LPIPS~\cite{zhang2018lpips}, DISTS~\cite{ding2020dists}, CLIP-IQA~\cite{wang2023clipiqa}, MUSIQ~\cite{ke2021musiq}, MANIQA~\cite{yang2022maniqa}, and NIQE~\cite{zhang2015niqe}. A deployment-aware submission process, including FP32 result packaging and strict reproducibility assets, makes the reported scores can be traced to concrete checkpoints.

\noindent\textbf{Implementation Details.}
The team starts from the challenge baseline checkpoint (CompTuneB @ iter 12000), and then performs multi-stage tuning. The first stage is an IQA-guided short fine-tune (batch size $2$ per GPU, patch size $224$, total $6000$ iterations, learning rate $1\times10^{-5}$ with milestones at $[2000,4000]$), using an L1 + IQA composite loss. The second stage is score-oriented fine-tuning (batch size $1$, patch size $192$, total $2000$ iterations, learning rate $2\times10^{-6}$ with milestones at $[800,1400]$), where GAN is disabled for stability and the objective emphasizes LPIPS/DISTS with lighter CLIP-IQA/MANIQA/MUSIQ constraints.

For model selection, VIPSL uses a local full-score proxy following the official metric set on RealSR(V3) and DRealSR. During the testing-phase submission, inference is run in FP32 mode with pre-padding ($16$), and the outputs are saved as lossless PNG format. Their final checkpoint is selected at the iteration $1000$.

\begin{figure*}[t]
\centering
\includegraphics[width=0.92\textwidth]{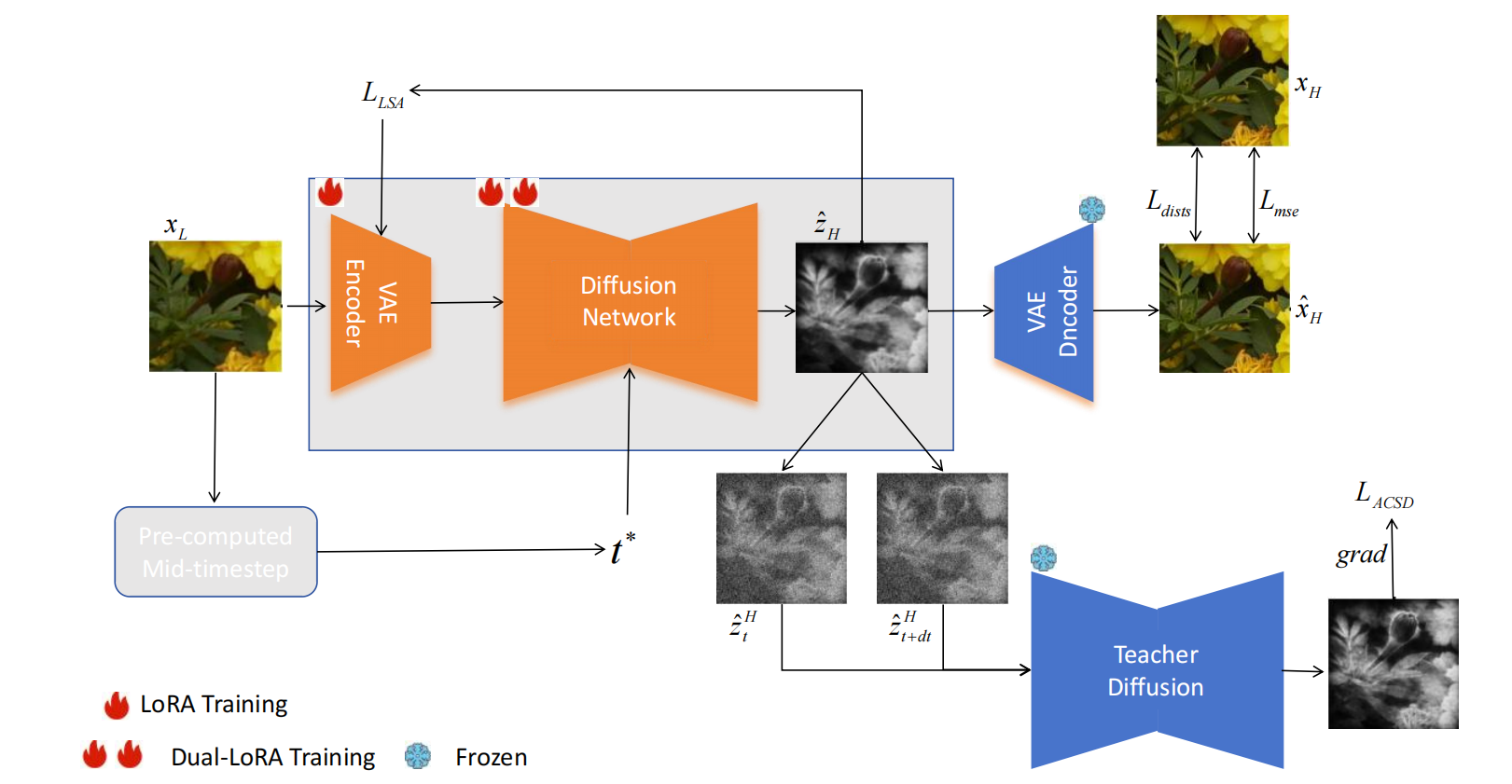}
\caption{\textbf{Team TODSR}}
\label{fig:team03_structure}
\end{figure*}
\subsection{Antman}
\textbf{Description.}
Antman uses RRDBNet from ESRGAN~\cite{wang2018esrgan} as the main generator and initializes from Real-ESRGAN x4plus~\cite{wang2021realesrgan}. As shown in Fig.~\ref{fig:team06_structure}, the architecture remains simple and fully convolutional, which keeps optimization stable and avoids extra deployment overhead. The method does not rely on complicated multi-branch structures; instead, it improves perceptual quality by redesigning the optimization objective.

Compared with purely reconstruction-oriented training, Antman introduces perceptual score guidance to better match challenge metrics. This includes no-reference quality predictors (MANIQA~\cite{yang2022maniqa}, MUSIQ~\cite{ke2021musiq}, CLIP-IQA~\cite{wang2023clipiqa}) and full-reference perceptual distances (LPIPS~\cite{zhang2018lpips}, DISTS~\cite{ding2020dists}). The resulting model keeps the robustness of RRDB-style restoration while moving outputs toward stronger perceptual realism.

\noindent\textbf{Implementation Details.}
The method keeps the standard Real-ESRGAN x4plus configuration: $3$ input/output channels, $64$ feature channels, $32$ growth channels, and $23$ RRDB blocks. Fine-tuning is carried out on the DIV2K unknown-degradation training split~\cite{agustsson2017div2k} with random crop and flip augmentation. HR patch size is $192\times192$ and mini-batch size is $4$. Optimization uses Adam for $10$K iterations, starting from learning rate $5\times10^{-5}$ and halving at $9000$ iterations, with EMA enabled. The training objective combines L1 reconstruction and weighted score-guided loss terms. The score terms are set to increase MANIQA/MUSIQ/CLIP-IQA and reduce LPIPS/DISTS, with scores computed on clamped outputs in $[0,1]$ for stability. This compact setup achieves strong perceptual quality without changing the core RRDB architecture.

\subsection{SamsungAICamera}
\textbf{Description.}
SamsungAICamera proposes an efficient diffusion-based SR framework by distilling a strong one-step diffusion teacher (OSEDiff~\cite{wu2024osediff}) into a compact student. The student consists of a pruned U-Net and a lightweight tiny autoencoder decoder (TAESD~\cite{bohan2023taesd}), and then adds a shallow Detail Enhancement Module (DEM) between denoising and decoding, as shown in Fig.~\ref{fig:team11_structure}. It inherits diffusion priors while reducing compute compared with heavier diffusion SR systems such as ADCSR~\cite{chen2025adversarial}.

The main motivation is to preserve perceptual richness while controlling runtime overhead on device-oriented settings. The DEM is intentionally shallow and only refines latent details before decoding, avoiding full-network retraining at the last stage. This yields a practical balance between quality improvement and latency budget.

\noindent\textbf{Implementation Details.}
Training is divided into three stages. Stage 1 performs distillation from a large teacher to the compact student using distillation and adversarial objectives. Stage 2 refines perceptual fidelity with L1, LPIPS~\cite{zhang2018lpips}, and DISTS~\cite{ding2020dists}, while simplifying the degradation process and introducing blur-like corruption for better robustness. Stage 3 freezes the pruned U-Net and decoder and updates DEM only, with additional IQA-oriented constraints using CLIP-IQA~\cite{wang2023clipiqa}, MANIQA~\cite{yang2022maniqa}, and MUSIQ~\cite{ke2021musiq}. The training data combine DIV2K~\cite{agustsson2017div2k}, LSDIR~\cite{li2023lsdir}, FFHQ~\cite{karras2019ffhq}, RealSRv4, and DRealSR. The factsheet reports a weighted perceptual objective with weights: L1 ($0.117$), LPIPS ($0.115$), DISTS ($0.244$), CLIP ($0.0326$), MANIQA ($0.119$), and MUSIQ ($0.079$). Learning rates are $1\times10^{-4}$ for stages 1--2 and $1\times10^{-5}$ for stage 3. For final inference, tiled processing is used (tile size $96$, overlap $32$). Their reported runtime decomposition is $720$ ms (U-Net), $30$ ms (decoder), and $8$ ms (DEM) on a Dimensity 9500 FP32 platform, with final factsheet score highlights of CLIP-IQA $=0.91$ and MUSIQ $=75.65$.

\begin{figure*}[t]
\centering
\includegraphics[width=0.9\textwidth]{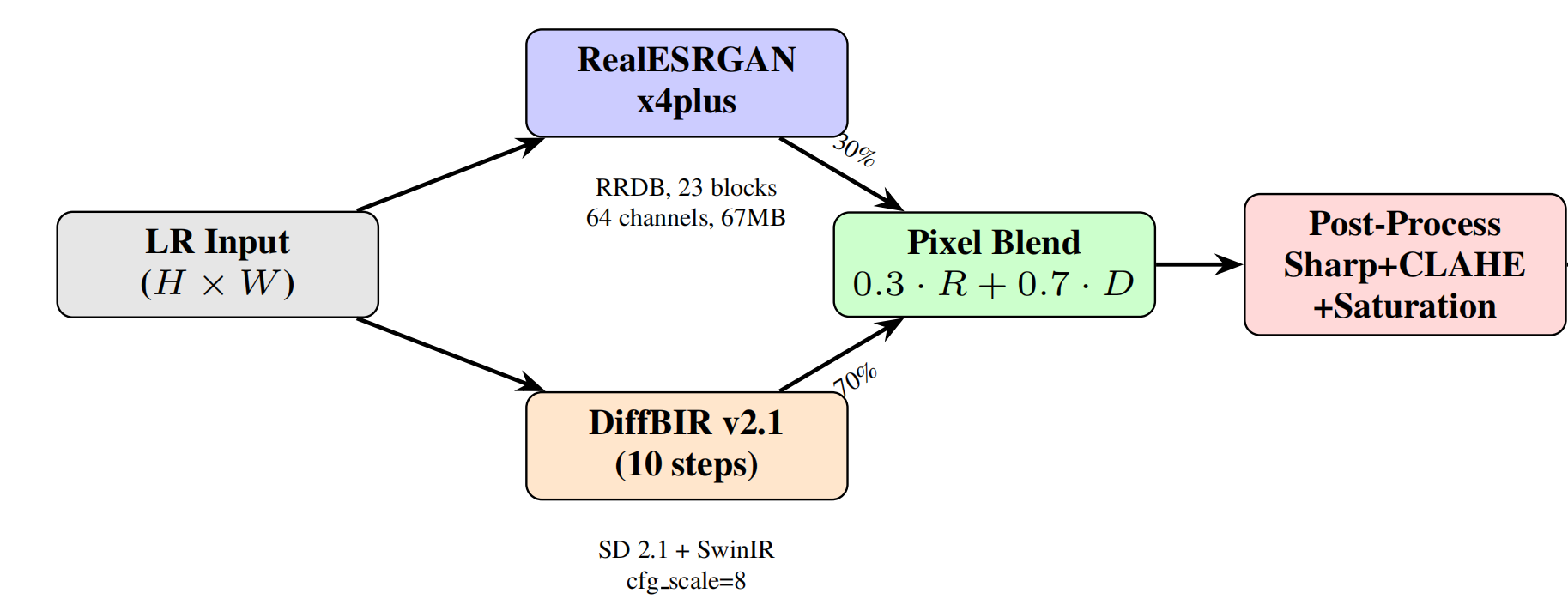}
\caption{\textbf{Team YuFans}}
\label{fig:team15_structure}
\end{figure*}

\subsection{TODSR}
\textbf{Description.}
TODSR is a one-step diffusion SR framework designed around better timestep utilization in latent diffusion~\cite{rombach2022ldm}. As shown in Fig.~\ref{fig:team03_structure}, the method proposes latent statistical-level alignment (LSA) to align low-quality latent features with diffusion latent states, and asynchronous conditional score distillation (ACSD) to exploit timestep-aware guidance during training. The design is motivated by the observation that fixed or weakly-used timestep conditions in one-step pipelines limit the use of generative priors.

The method is closely related to recent one-step real SR diffusion advances~\cite{wu2024osediff} and extends ideas from PiSA-SR~\cite{sun2025pisasr} with a dual-LoRA optimization strategy~\cite{hu2022lora}. It further incorporates insights from OMGSR~\cite{wu2025omgsr} and FaithDiff~\cite{chen2025faithdiff}, emphasizing that latent alignment and semantic guidance should be jointly coordinated to avoid unstable over-enhancement.

\noindent\textbf{Implementation Details.}
In the first training stage, TODSR fine-tunes the VAE encoder and diffusion network jointly, using pixel-level objectives and LSA to match latent statistics at a precomputed intermediate timestep ($t^\ast=273$ in the factsheet). In the second stage, the first LoRA branch is frozen and a second LoRA branch is optimized with ACSD and DISTS supervision~\cite{ding2020dists}, which strengthens semantic details and perceptual realism. The team then performs additional challenge-oriented fine-tuning with full-reference and no-reference IQA losses.

The implementation uses SD2.1 as base diffusion prior~\cite{rombach2022ldm}, AdamW optimization~\cite{loshchilov2019adamw}, and mixed high-quality datasets including DIV2K~\cite{agustsson2017div2k}, DIV8K~\cite{gu2019div8k}, and LSDIR~\cite{li2023lsdir}, with degradation generation following PASD settings~\cite{yang2023pasd}. Training is run on one RTX4060 GPU, with $512\times512$ patches and batch size $16$; pix-LoRA is trained for $4500$ iterations and sem-LoRA for $16000$ iterations at initial learning rate $5\times10^{-5}$. The team reports a best testing-phase score of $4.2824$ and an estimated speedup ratio of $2.8671$ over OSEDiff in their benchmark setting.

\subsection{YuFans}
\textbf{Description.}
YuFans proposes a practical ensemble called DiffBIR-RealESRGAN Perceptual Blend. The framework combines a diffusion-based blind restoration model, DiffBIR v2.1~\cite{lin2024diffbir}, with Real-ESRGAN~\cite{wang2021realesrgan}. As shown in Fig.~\ref{fig:team15_structure}, the two outputs are fused by a fixed pixel-level ratio, and then a lightweight post-processing module is applied. The approach is simple but effective for perceptual optimization. The core rationale is complementarity: diffusion outputs are often richer in perceptual detail but may introduce unstable artifacts, while GAN outputs are structurally cleaner but less vivid. By blending both branches, the method preserves high-level realism from diffusion while injecting structural stability from GAN restoration.

\noindent\textbf{Implementation Details.} The final pipeline has four stages: Real-ESRGAN restoration, DiffBIR restoration, weighted blending, and perceptual post-processing. In the factsheet implementation, DiffBIR is run with 10 denoising steps and classifier-free guidance scale $8$, and the blend is $0.7\cdot I_{\text{DiffBIR}}+0.3\cdot I_{\text{RealESRGAN}}$. The post-processing stage applies unsharp masking ($\sigma=3$, strength $2.0$), CLAHE (clip limit $4.0$, tile $8\times8$), and saturation boost (factor $1.2$), producing additional gains on CLIP-IQA and MUSIQ.

A key advantage is that no additional model retraining is required. Both backbones are used with pretrained weights, and only blending/post-processing hyperparameters are tuned on the validation set. The team reports systematic blend-ratio ablations and notes that pure diffusion outputs may exceed platform time limits, while blending retains most perceptual gains with better practical robustness.

\subsection{IMAG2006}
\begin{figure*}[t]
    \centering
    \includegraphics[width=0.9\textwidth]{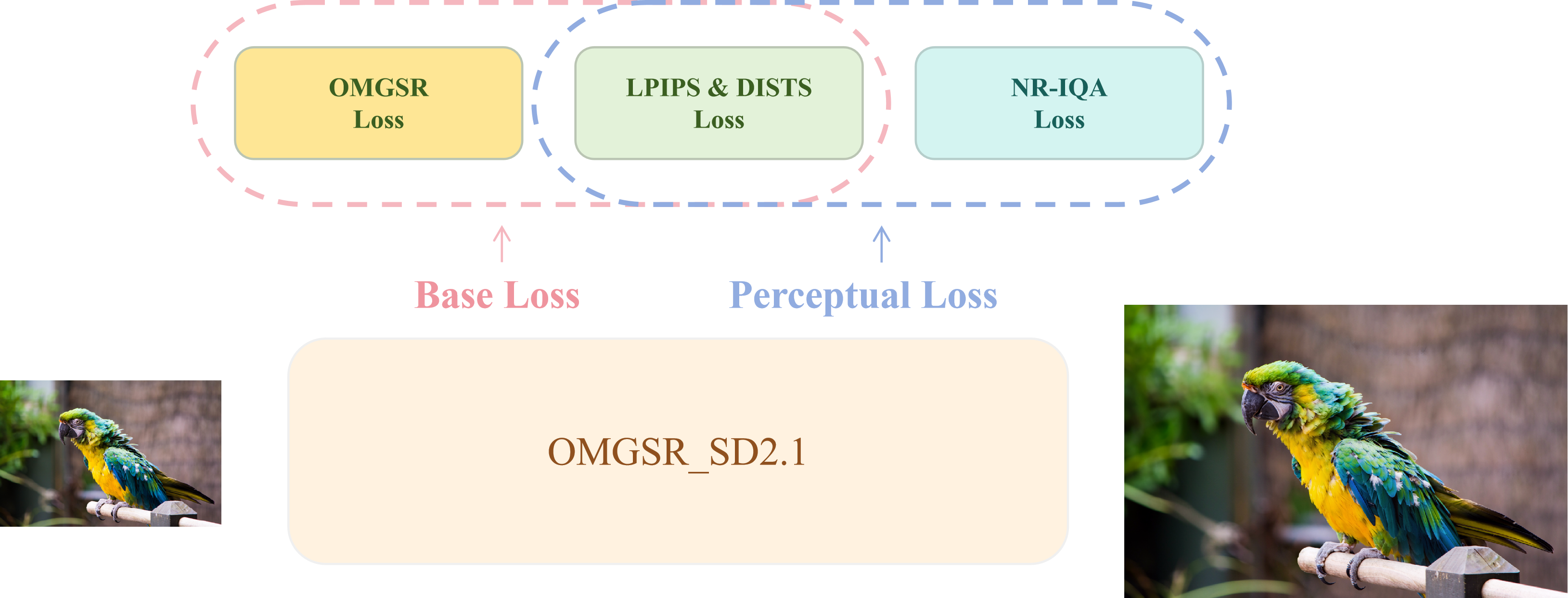}
    \caption{\textbf{Team IMAG2006}}
    \label{fig:team_imag2006}
\end{figure*}

\textbf{Description.} The proposed solution is shown in Fig.~\ref{fig:team_imag2006}. The method of Team IMAG2006 is mainly built upon OMGSR~\cite{wu2025omgsr}, a one-step super-resolution approach based on diffusion models. It achieves low-quality (LQ) image injection for restoration by only adjusting the diffusion timestep corresponding to the LQ input.

During training, Team IMAG2006 replace the original DINOv3-based DISTS loss in OMGSR with the LPIPS loss~\cite{zhang2018lpips}, denoted as $\mathcal{L}_{\mathrm{LPIPS}}$, and the DISTS loss~\cite{ding2020dists}, denoted as $\mathcal{L}_{\mathrm{DISTS}}$, to provide richer perceptual supervision and improve the perceptual relevance of the training objective.

\begin{equation}
\begin{aligned}
\mathcal{L}_{\mathrm{base}}
=&\;
\lambda_{1} \mathcal{L}_{\mathrm{LRR}}
+
\lambda_{2} \mathcal{L}_{\mathrm{L1}}
+
\lambda_{3} \mathcal{L}_{\mathrm{GAN}} \\
&+
\lambda_{4} \mathcal{L}_{\mathrm{LPIPS}}
+
\lambda_{5} \mathcal{L}_{\mathrm{DISTS}},
\end{aligned}
\end{equation}
where $\mathcal{L}_{\mathrm{LRR}}$ is used to align the LQ input at the specified timestep with the noised HQ target, $\mathcal{L}_{\mathrm{L1}}$ is introduced to preserve pixel-wise fidelity, and $\mathcal{L}_{\mathrm{GAN}}$ is employed to enhance visual realism.

Since OMGSR is a one-step diffusion method based on a pre-trained model, it can preserve the rich prior knowledge of the pre-trained backbone. To guide the model toward producing images with stronger visual perception, Team IMAG2006 follows~\cite{zhang2025augmenting} and incorporate no-reference image quality assessment (NR-IQA) losses~\cite{ke2021musiq,wang2023clipiqa,yang2022maniqa} as additional supervision during training.

The corresponding training objective is given by:
\begin{equation}
\begin{aligned}
\mathcal{L}_{\mathrm{NR-IQA}}
=&\;
\lambda_{6} \mathcal{L}_{\mathrm{MUSIQ}}
+
\lambda_{7} \mathcal{L}_{\mathrm{CLIPIQA}} \\
&+
\lambda_{8} \mathcal{L}_{\mathrm{MANIQA}},
\end{aligned}
\end{equation}
where $\mathcal{L}_{\mathrm{MUSIQ}}$, $\mathcal{L}_{\mathrm{CLIPIQA}}$, and $\mathcal{L}_{\mathrm{MANIQA}}$ are all differentiable and therefore can be used as perceptual supervision signals. For metrics where higher scores indicate better perceptual quality, Team IMAG2006 first normalizes the scores to $[0,1]$ and then transforms them into minimization objectives by using $1-s$, where $s$ is the normalized score. 

The total training loss is formulated as:
\begin{equation}
\begin{aligned}
\mathcal{L}_{\mathrm{total}}
=&\;
\mathcal{L}_{\mathrm{base}}
+
\mathcal{L}_{\mathrm{NR-IQA}}
\end{aligned}
\end{equation}
\textbf{Implementation Details.} Team IMAG2006 trains the model using the AdamW optimizer with $\beta_1=0.9$, $\beta_2=0.999$, and a learning rate of $5\times10^{-5}$. The ground-truth images are cropped into $512\times512$ patches. Training is conducted for 6{,}000 iterations with a batch size of 4. The loss weights are set as $\lambda_1=5$, $\lambda_2=0.5$, $\lambda_3=0.5$, $\lambda_4=3$, $\lambda_5=3$, $\lambda_6=0.05$, $\lambda_7=2$, and $\lambda_8=1$. The VAE rank is set to 16, while the UNet rank is set to 32. Following FaithDiff~\cite{chen2025faithdiff}, Team IMAG2006 uses the same training datasets, including DIV2K, LSDIR, and other high-quality datasets.

\section{Methods of the Remaining Teams}
All participating teams contributed meaningful method designs and practical training strategies for the NTIRE 2026 Mobile Real-World Image Super-Resolution challenge. Due to space limitations, we focus the main manuscript on the top-ranked methods in Sec.~\ref{sec:methods} and summarize the remaining teams here in a concise manner. Their solutions cover a broad spectrum of ideas, including efficient CNN/Transformer restoration backbones, diffusion prior based enhancement, perceptual-loss reweighting, data degradation simulation, and multi-model fusion.

Although these methods are not individually detailed in the main text, they provide valuable technical diversity and reflect different tradeoffs among perceptual quality, robustness, and inference efficiency. We include their complete method descriptions, implementation settings, and additional ablation details in the supplementary materials to ensure fair visibility of all valid submissions.

\section*{Acknowledgments}
Transsion Holdings is a co-organizer of this work.
This work is supported by the National Natural Science Foundation of China (62501386, 625B2116, 625B1025), CCF-Tencent Rhino-Bird Open Research Fund. This work is also sponsored by Al Hundred Schools Program and is carried out using the Ascend AI technology stack.
This work was partially supported by the Humboldt Foundation. We thank the NTIRE 2026 sponsors: OPPO, Kuaishou, and the University of Wurzburg (Computer Vision Lab).

{\small
\bibliographystyle{ieeenat_fullname}
\bibliography{main}
}

\end{document}